\documentclass[preprint,12pt, 3p]{elsarticle}

\usepackage{array}
\usepackage{multirow}
\usepackage{amsmath, amsthm, amssymb}
\usepackage{listings}

\usepackage{algpseudocode}
\usepackage{rotating}
\usepackage{graphicx}
\usepackage{booktabs}
\usepackage{adjustbox}
\usepackage{subcaption}
\usepackage{xcolor}
\usepackage{float}
\usepackage{booktabs}
\usepackage[ruled,vlined]{algorithm2e}
\usepackage{hyperref}
\usepackage{enumerate}

\newtheorem{defi}{Definition}
\newtheorem{ex}{Example}

\begin{document}
\begin{frontmatter}

\title{PRecG: Legal Precedent Retrieval with Graph Neural Networks and Rhetorical Role Segmentation}

\author[1]{Devanshu Verma}
\ead{dverma@cs.du.ac.in}
    
\author[1]{Vasudha Bhatnagar}
\ead{vbhatnagar@cs.du.ac.in}
  
\author[1]{Vikas Kumar\corref{cor1}}
\ead{vikas@cs.du.ac.in}

\author[2]{Balaji~Ganesan}
\ead{bganesa1@in.ibm.com}

\address[1]{University of Delhi, Delhi, India}    
\address[2]{IBM Research, Bengaluru, India}  

\cortext[cor1]{Corresponding author}




\begin{abstract}
Legal precedent retrieval is a fundamental task in legal case preparation, planning, litigation strategy, and legal research. Current approaches for automatic precedent retrieval map legal documents to a low-dimensional semantic space and compute similarity based on the proximity of their representations. These approaches treat legal documents as monolithic texts, ignoring the rhetorical organization of the legal technicalities. Ergo, they overlook nuanced legal meanings and fail to distinguish the contextual significance of legal entities and concepts that vary based on their rhetorical roles within the document. 

To address this insufficiency, we propose the PRecG pipeline that computes the similarity between pairs of legal judgments by hierarchically learning their representations. The process begins by decomposing each document into distinct semantic units (segments) based on the rhetorical roles of sentences. For each rhetorical segment, a knowledge graph is constructed to capture the legal entities and their relationships within the segment. Contextual representations of the entities are then learned and aggregated to derive segment-level embeddings. These embeddings are further integrated to produce a unified document-level representation, and finally, the semantic similarity between a pair of documents is computed.
We validate the performance of the proposed approach through extensive experiments on a benchmark Indian legal dataset, comparing it against state-of-the-art baselines to demonstrate its effectiveness. 

\end{abstract}




\end{frontmatter}

\section{Introduction}

Legal systems aim to ensure that human actions within society are kept in order and control. They form an integral part of a nation's culture, civilization, history, and the everyday life of its people. 
Different countries around the world have various jurisdictions, including \textit{civil law}, \textit{common law}, \textit{customary law}, \textit{religious law}, and \textit{mixed law}. 
India, in particular, operates under a mixed legal system, with the common law having a prominent role. This system adheres to the concept of stare decisis,
 which accords equal importance to codified statutes and prior case judgments \cite{landes2013legal}. 
The fundamental principle is that cases with similar facts and situations should be decided consistently, in accordance with the judgments of prior similar cases or precedents.

\begin{figure} 
    \centering
    \includegraphics[width=.5\linewidth]{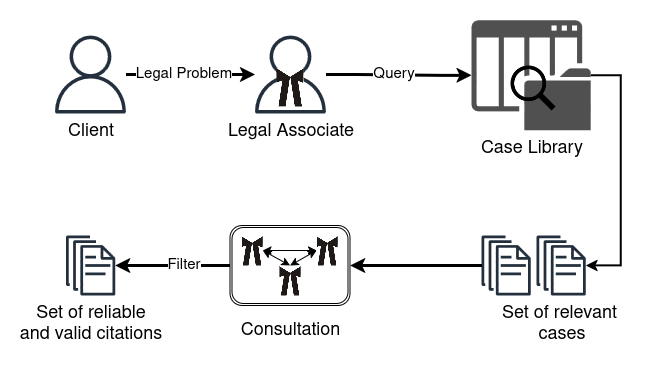}
    \caption{Manual procedure for finding valid and reliable citations.}
    \label{fig:fig_1}
\end{figure}

Precedent cases are integral to the common law system and serve as the basis for judicial decision-making \cite{gerhardt2011power}. Drawing upon prior judgments to deliver verdicts promotes consistency, transparency, and fairness in legal proceedings~\cite{landes2013legal}. 
Therefore, ensuring the relevance and reliability of precedents is essential and must be strongly adhered.
Typically, the process of precedent selection begins with a keyword-based search from an e-repository of legal cases, and a trained professional manually selects a set of relevant keywords as a query to retrieve a candidate set of cases. The search engine retrieves potentially relevant documents from the repository, and the legal experts carefully review and rank the retrieved cases. Often, a consultation process results in the final selection of precedent cases.  This process, depicted in Figure~\ref{fig:fig_1}, is inherently time-consuming and prone to limitations arising from human expertise in keyword selection, manual indexing, and ranking of retrieved documents.  

Automating the task of retrieving \textit{precedent} case documents can significantly aid legal professionals at various stages of adjudication and ruling. For lawyers, it helps establish a strong foundation for a new case by efficiently identifying relevant \textit{precedents} from a large corpus of judgments. For judges, automation of \textit{precedent} retrieval assists in delivering well-grounded verdicts by ensuring consistency with each relevant ruling. Automation of the \textit{precedent} retrieval task also helps mitigate human bias in precedent selection, thereby enhancing the consistency and reliability of judicial outcomes.

Early methods for automatic precedent retrieval largely relied on statistical and machine learning approaches using bag-of-words representation~\cite{kumar2013finding, o2016analysis}. These approaches relied on surface-level lexical features and required extensive manual feature engineering. While they offered initial promise in retrieving relevant precedents, their performance was often limited by their inability to capture deeper semantic or contextual nuances in legal language~\cite{aletras2016predicting}.

Recent advances in Natural Language Processing (NLP) and Deep Learning (DL) have contributed to more accurate and efficient approaches for automatic precedent retrieval. Given a query and a document corpus, these methods select relevant precedents from the corpus by embedding the query and corpus documents in the same latent space and computing similarity between them~\cite{paul2022pretraining, hong2020lfesm, bi2022heterogeneous}.
Existing methods learn the \textit{unified} representation of the documents in a latent semantic space and capture the contextual semantics of legal concepts. However, these methods typically process documents as a monolith for learning the representation, which makes them prone to inaccuracies because of the length and linguistic complexity of legal texts. Moreover, legal terminology often conveys different meanings, depending on the section in which it appears. For example, the term \textit{negligence} serves distinct functions when appearing in different contexts. In the \textit{factual background} section, the plaintiff may describe the defendant’s failure to maintain safe premises as \textit{negligence}, referring to a specific act. However, in the \textit{legal reasoning} section, the court might treat it as a broader legal standard, requiring proof that the defendant breached a duty of care owed to the plaintiff. These contextual differences can lead to suboptimal performance of the automatic precedent retrieval systems in two key scenarios: $(i)$ when the query and precedent refer to similar terms but in different contexts (i.e., semantic roles), and $(ii)$ when the representation of a lengthy and structurally complex legal document dilutes locally relevant concepts, thereby impairing retrieval accuracy.

In this paper, we address the limitations of existing automatic precedent retrieval models, which view legal documents as monolithic texts \cite{bi2022heterogeneous, kumar2013finding, o2016analysis}. Our approach models the similarity between a pair of case documents by first learning their representations at a finer granularity and aggregating them to construct a document-level embedding. Specifically, we begin by automatically decomposing each document into \textit{segments} based on the rhetorical roles of sentences. For each segment, we construct a segment-level knowledge graph and learn the representation of legal entities and their contextual dependencies using a Graph Neural Network. These learned representations of the entities are then aggregated using attention-based pooling to construct segment-level embeddings. The segment-level embeddings are aggregated using a transformer-based mechanism to form a \textit{unified} representation of the case document, and the similarity score between the pair of documents is computed. During training, any error in the predicted similarity score between a pair of documents is corrected by updating the model parameters. At inference time, the embedding of a given query text is obtained from the learned model, and its similarity with the candidate documents is computed. The top-$k$ most similar documents are identified as the most relevant precedents.

We summarize the contributions of the paper below.
\begin{enumerate} [i.]
    \item We curate a legal ontology tailored for the Indian legal domain (Section \ref{sec:legont}).
     \item We propose an end-to-end learning framework that first decomposes the legal document into semantic segments based on the rhetorical roles of the sentences, learns the representation of each segment, and then aggregates them using an attention-based aggregation mechanism into unified document-level embeddings (Section \ref{sec:methodology}).
    \item We leverage a Large Language Model to automatically extract legal entities and their relationships for automatic construction of segment-level knowledge graphs (Section \ref{subsec:egcons}).
    \item We use a Graph Neural Network to enhance the representation of legal entities within each segment-level knowledge graph (Section \ref{subsec:replearn}).
    \item We describe the experimental setup in Section \ref{sec:experi} and report the results in Section  \ref{sec:result}.
\end{enumerate}

Section \ref{sec:relwork} describes the related work, followed by the background and preliminaries in Section \ref{sec:probstat}.
Section \ref{sec:concl} presents the conclusion and scope for future work.

\section{Related Work}
\label{sec:relwork}

Legal text representation is the foundational step in assessing case similarities. We review the existing methods that have been used in the automatic precedent retrieval task. We also relate our methodology to existing works that use a knowledge graph for legal text analytics.

\subsection{Legal Text Representation}

A key challenge in precedent retrieval is to represent a legal judgment in a machine-readable form that preserves the semantics, context, and domain-specific legal intricacies. 
Early approaches treat legal documents as a bag-of-words, and rely on statistical methods such as TF-IDF and topic modeling to represent them \cite{kumar2013finding, o2016analysis}. However, these methods are inadequate for representing the rich semantics and contextual depth of legal texts~\cite{aletras2016predicting}.

Neural methods such as Word2Vec \cite{xiachu2019wv} and GloVe \cite{pennington2014glove} improve upon statistical methods by capturing semantics. Transformer models like BERT \cite{devlin2019bert} further advance contextual representation. However, these general-purpose embeddings lack the specialized legal knowledge required to faithfully represent the complex and nuanced information present in legal documents.

To address this gap, domain-enriched models have been proposed, which incorporate legal knowledge for legal document representation. Hong et al. \cite{hong2020lfesm} introduce the Legal Feature Enhanced Semantic Matching Network (LFESM), which uses regex-based legal feature extraction to improve semantic matching. Although this approach benefits from integrating domain cues, the range and depth of encoded legal features remain limited. Recently, Bi et al. \cite{bi2022heterogeneous} proposed L-HetGRL, a heterogeneous graph-based representation model for Chinese legal documents that integrates linguistic and structural information through legal knowledge graphs.

Despite the advances in legal knowledge-enriched contextual representation, existing approaches represent a legal case using its entire text, disregarding its internal structure and rhetorical organization. This is particularly problematic for Indian legal documents, which are typically verbose, repetitive, and lack consistent structure. These characteristics make it difficult to algorithmically distinguish legally significant content from generic text \cite{bhattacharya2019identification}. As a result, \textit{the above-mentioned approaches} risk missing contextually significant information, which reduces their effectiveness in precedent retrieval.

\subsection{Knowledge Graphs in the Legal Domain}

Knowledge graphs (KG) serve as structured representations of information consisting of entities, relationships, and semantic descriptions. Entities represent real-world objects or abstract concepts, while relationships define how these entities are interconnected.

In legal text analytics, KGs offer a principled way to represent the inherently relational knowledge of legal texts. They help structure legal documents by capturing intricate connections among entities such as statutes, judicial principles, case facts, and legal actors. This structured representation enhances contextual understanding and the interpretability of the case documents and has proven valuable in applications like legal question-answering \cite{huang2019knowledge}, legal reasoning \cite{singhal2012introducing}, etc. 

Legal texts are often complex, lengthy, and rich in domain-specific jargon, which makes knowledge graph (KG) construction a non-trivial task. Building effective legal KGs requires sophisticated pipelines to accurately identify and link important legal entities and relationships while filtering out redundant or irrelevant content. Addressing these challenges is crucial to fully realizing the potential of KG-based methods in legal applications.

\section{Problem Description and Key Concepts}
\label{sec:probstat}

The Legal Precedent Retrieval (LPR) task aims to find which among the given set of prior cases should be noticed or considered relevant to a given query case \cite{raman2008lcr}. The objective is achieved by analyzing the set of cases and identifying those that share facts, legal principles, or contextual elements with a given query case. The query case, denoted by $Q$, refers to the factual and contextual description of a legal situation for which the legal expert is searching for relevant precedent cases. The set of prior cases, denoted by $\mathcal{J} = \{J_1, J_2, \cdots, J_N\}$, is curated by legal experts either manually or using a keyword-based search, or both. The goal of the LPR task is to facilitate the retrieval of $\mathcal{P} \subseteq \mathcal{J}$,  an ordered subset of cases most relevant to  $Q$. The cases in $\mathcal{P}$ are ranked based on the relevance score $\mathcal{\rho}$, which quantifies their jurisprudential significance and contextual alignment with the query. 

\subsection{Key Concepts and Notations}

In this section, we introduce the fundamental concepts and components that form the basis of the proposed methodology, establishing a clear and precise foundation for our approach.

\begin{enumerate}[i.]
\item \textbf{Legal ontology:}
A legal ontology represents legal cognition as a structured set of concepts and relationships within the legal domain. It defines legal entities, along with their types and attributes, and the relations between them, along with the relation types. Examples of legal entity types include \textit{individuals, corporations, government agencies, courts, and contracts, etc.} Examples of legal relation types between legal entities include \textit{employs}, \textit{regulates}, and \textit{files\_case\_against}, etc.

\begin{defi}
    Given a set of legal entity types $T_E$ and a set of legal relationship types $T_R$, a legal ontology $\mathcal{O}$ is a collection of tuples $\{(t_i, t_j, R_{ij})\}$, where $t_i, t_j (\in T_E) $ are the head and tail concepts respectively, and $R_{ij} \subset T_R$ represents the set of valid legal relationship types between $t_i$ and $t_j$.
\end{defi}
\begin{ex}
  Let $T_E$ =  \{\textit{Person, Contract, Organization}\}  and $T_R$ = \{\textit{enters\_into}, \textit{employs}, \textit{represents}\}.  Then the tuple $\bigl( \text{\textit{Court}, \textit{Judgment}, \textit{\{issues\}}}\bigr) $in the ontology represents a "court issues a judgment".   
\end{ex}
\item \textbf{Semantic segmentation of a judgment:}
Semantic segmentation divides a case document into distinct functional units, thereby imposing a semantic structure on its content. It ensures that documents are broken into meaningful components that accurately reflect their legal functions and roles. In the context of the Indian judicial system, semantic segmentation is crucial for legal text analysis due to the absence of a standard format for writing judgments and the inherent diversity in the structure of case documents from Indian courts \cite{bhattacharya2019identification}. 
\begin{defi}
     \textbf{Semantic segmentation} of the case document $J_i$ is the process of dividing it into $k$ pre-specified distinct segments          
     $\{S_i^1,~ S_i^2,~ S_i^3,~ ...,~ S_i^k\}$, where segment $S_i^p$ contains sentences that correspond to a specific semantic role ``$p$" in $J_i$.
\end{defi}
\item \textbf{Legal knowledge graph:} 
The Legal Knowledge Graph (LKG) is an instantiation of the legal ontology, representing concrete instances of legal entity types and their relationships for legal document(s). The LKG of a case document is useful for understanding the dynamics of legal proceedings, the roles of different entities, predicting case outcomes, and identifying potential conflicts of interest. 
It is inherently heterogeneous, encompassing multiple entity types and relationships that capture the complexities of the legal system.
\begin{defi} 
A legal knowledge graph for a case document $J$ is defined as $\mathcal{G} = G(\mathcal{V}, \mathcal{E}) $, where $\mathcal{V}$ is the set of legal entities and $\mathcal{E}$ is the set of relations between them. Each entity $v (\in \mathcal{V})$ has a type $t \in T_E$, and each relation $(u, v) \in \mathcal{E} $ corresponds to a relation type $r \in R_{ij}$ between entities $u$ and $v$ of types $t_i$ and $t_j$, respectively. 
\end{defi}
\item \textbf{Vector representation:}
To enable computational analysis, the entire case document, including legal entities, relations, and semantic segments, is embedded in a continuous vector space $\mathbb{R}^d$. The document embedding captures the latent semantics and legal intricacies of legal concepts, enabling the computation of document similarity. Table~\ref{tab:vecNotation} summarizes the notation for the vectorized representation of the components used in the PRecG framework.

\begin{table}[h]
\centering
\hfill
\caption{Vector notation for legal elements.}
\label{tab:vecNotation}
\begin{tabular}{ll}
\toprule
\textbf{Symbol} & \textbf{Description} \\
\midrule
$\mathbf{e}$ & Embedding of a legal entity $e$ \\
$\mathbf{r}$ & Embedding of a legal relation $r$ \\
$\mathbf{s}_i^p$ & Embedding of the $p$-th segment $S_i^p$ of document $J_i$ \\
$\mathbf{j}_i$ & Embedding of the $j$-th legal document $J_i$ \\
\bottomrule
\end{tabular}
\end{table}

\item \textbf{Graph representation learning: }

Graph Representation Learning (GRL) is an effective approach for transforming raw graphs into low-dimensional embeddings while preserving both local and global contexts. The key to preserving contextual information lies in the joint consideration of node attributes, neighborhood characteristics, and relational patterns among entities using Graph Neural Networks (GNNs). GNNs follow an iterative message-passing framework, where each node updates its representation by aggregating information from its local neighborhood. Through successive iterations, these localized updates capture higher-order dependencies and help in obtaining contextualized representations.

\item \textbf{Inductive precedent retrieval:} 
Legal Precedent Retrieval (LPR) involves training a model on pairs of legal case documents, where each document is encoded into a representation that captures both structural and semantic features. During training, the model learns a transformation function that maps the document’s knowledge graph to a comprehensive document-level embedding. This follows the principle of inductive learning, as the learned transformation generalizes to unseen cases. Once trained, the transformation embeds any legal case document into a shared latent space.
At inference time, a knowledge graph is first constructed for the query document and passed through the trained transformation module to obtain its vector representation. This embedding is then compared with those of all documents in the corpus using similarity computation, and the top $k$ closest documents form the set $\mathcal{P}$ of relevant precedents.

\end{enumerate}

\section{Legal Ontology Creation}
\label{sec:legont}

The adoption of ontology in legal text analytics has attracted serious attention among researchers. Crucial for capturing legal intricacies in structured and machine-readable form, a comprehensive legal ontology is instrumental for optimal performance of the proposed automatic precedent retrieval model. The existing ontologies for the Indian legal context lack representational power due to their limited coverage of legal entities and relations \cite{jain2020nyon, dube2022indilegalont}. 

We address this gap by expanding the publicly available NyoN ontology \cite{jain2020nyon} on civil, criminal, and intellectual property law, while ensuring a comprehensive and structured representation of Indian legal knowledge. 
\begin{figure} 
    \centering
    \includegraphics[width=0.6\linewidth]{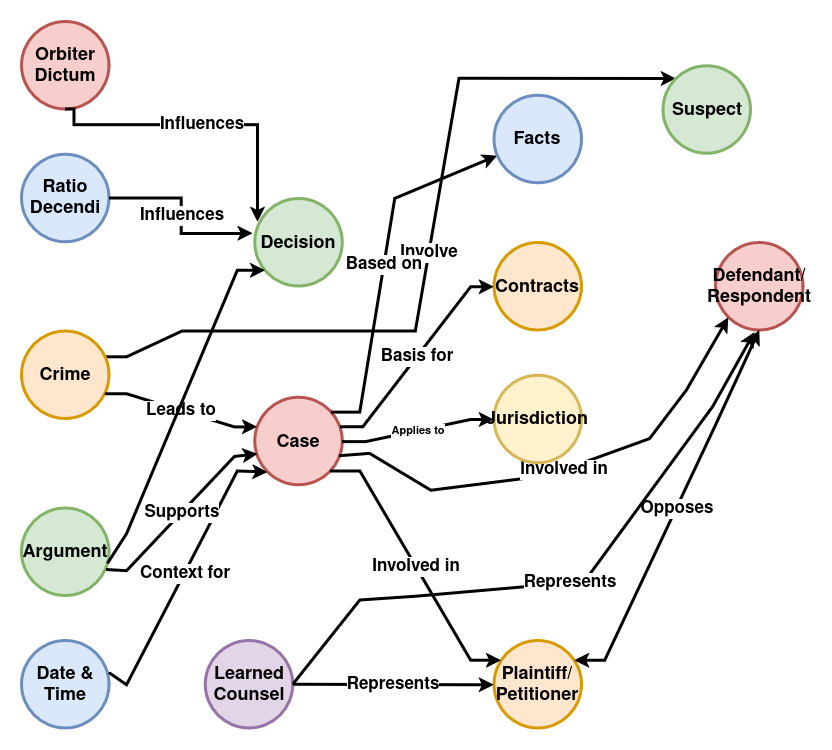}
    \caption{A snapshot from the constructed legal ontology.}
    \label{fig:l_ontology}
\end{figure}
We use the entities in the existing ontology as a seed set and curate a diverse set of legal entity types\footnote{We use the 1095 judgments scraped from the High Courts and Supreme Court of India for creating the ontology.} based on their relevance across different sub-domains of law. Subsequently, we populate the legal relation predicates using the Llama model with a few-shot prompting strategy\footnote{Due to paucity of space, we are unable to present the prompts in the manuscript. The details will be available as supplementary material after notification.}. To ensure reliability, we incorporate a human-in-the-loop validation step through an online platform where legal experts review, confirm, or refine the automatically generated relations. This guarantees that the ontology reflects not only syntactically plausible but also legally valid and contextually meaningful relations, making it suitable for downstream tasks of knowledge graph construction. A snapshot of the ontology is shown in Figure~\ref{fig:l_ontology}.

\begin{figure}
    \centering
    \includegraphics[width=0.7\linewidth]{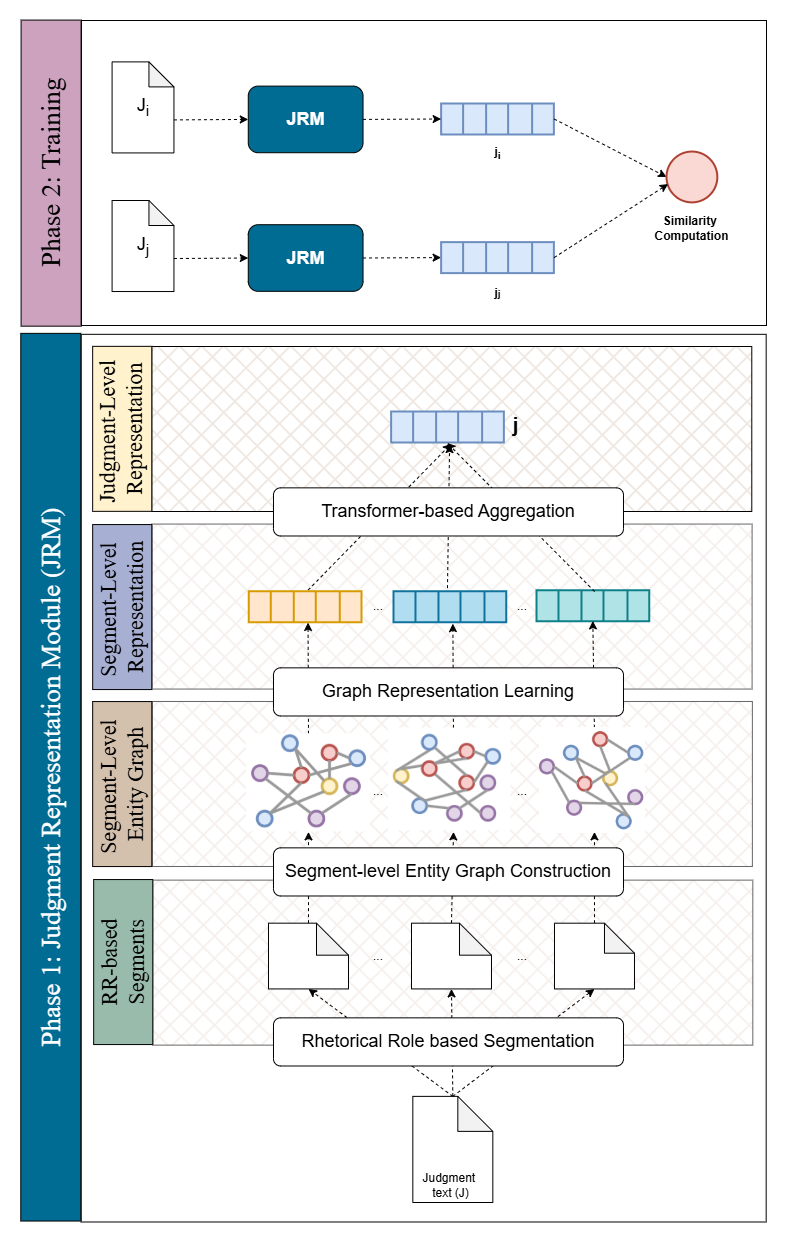}
    \caption{Technical details of the PRecG pipeline.}
    \label{fig:approach_pipeline}
\end{figure}

\section{Methodology}
\label{sec:methodology}
In this section, we present the details of the proposed PRecG model, which adopts a hierarchical strategy to model similarity between pairs of documents through five key stages:
(i) judgment segmentation,
(ii) construction of segment-level knowledge graphs,
(iii) contextual representation learning,
(iv) aggregation of segment-level representations, and
(v) similarity computation.
The pipeline is outlined schematically in Figure~\ref{fig:approach_pipeline}.  We delineate the details in the following subsections.

\subsection{Judgment Segmentation}
\label{subsec:segmen}

Indian judgments are typically lengthy and unstructured, without official formatting standards. This peculiarity makes information extraction from Indian judgments particularly challenging compared to judgments from other countries, as observed by Bhattacharya et al. \cite{bhattacharya2019identification}. Hence, it is imperative to partition the case proceedings into well-defined segments to capture their fine-grained structural and semantic organization. Traditional segmentation strategies such as fixed-length windows, discourse markers, or topical boundaries often prove inadequate because the precise meaning of legal concepts depends on their specific placement within a judgment. To address the nuances of legal language, we adopt rhetorical role-based segmentation of the judgment to effectively capture the underlying logical flow and argumentative structure.

Rhetorical roles such as \textit{facts, arguments, issues, precedents, statutes, rulings by lower court, and rulings by current court} are prevalent within legal case documents~\cite{bhattacharya2019identification}. Identifying text according to these roles helps in understanding the semantic landscape of a case document. Since manual annotation is impractical due to the length of judgments and the scale of legal corpora, we employ the sentence-level rhetorical role classifier HierBiLSTM-CRF~\cite{bhattacharya2019identification}, which classifies each sentence into one of the seven predefined rhetorical roles. Sentences with the same role are grouped together, thereby segmenting the case document $J_i$ into a set $\mathcal{S}_i = \{S^1_i, S^2_i, \dots, S^k_i\}$. Each segment $S^p_i$ consists of the sentences classified under role $p$.

\subsection{Segment-level Knowledge Graph Construction}
\label{subsec:egcons}

Given the rhetorical role-based segments of a case document, the next step is to obtain an effective representation of each segment. Legal reasoning relies on complex interactions between statutes, precedents, parties, issues, and more, which makes it essential to explicitly capture semantic dependencies within each segment. Traditional textual encoding methods, such as bag-of-words or sentence embeddings, fail to capture these structured interactions. Hence, we adopt a knowledge graph-based representation approach, in which segments are transformed into graphs with legal entities represented as nodes and their contextual relationships as edges. 
The knowledge graph for each segment is constructed in two steps, described below.

\begin{enumerate}[i)]

\begin{figure}[ht]
    \centering
    \includegraphics[width=0.7\linewidth]{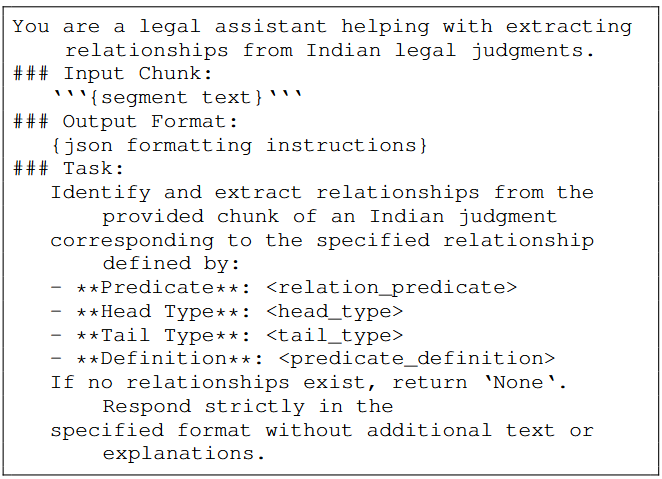}
    \caption{Prompt for triplet extraction from Indian Court Judgment.}
    \label{fig:ext_prompt}

\end{figure}

\item \textbf{Triplet extraction}:
We adopt a knowledge-based approach and leverage the advanced natural language understanding capabilities of large language models (LLMs)~\cite{xu2024large}. For entity and relation extraction, we make use of the \textit{Llama 3.1 8B model}\footnote{\url{https://huggingface.co/meta-llama/Llama-3.1-8B}}, an open-source LLM pretrained on general-domain corpora. 
To ensure that the extracted knowledge is aligned with domain-specific semantics and structurally consistent with the Indian judicial system, we leverage the created legal ontology (mentioned in Section \ref{sec:legont}).
We tailored the prompt (Figure~\ref{fig:ext_prompt}) to guide the LLM for extracting triples for the segment-level knowledge graph that conform to the predefined schema.

\item \textbf{Entity canonicalization}:

Triplet extraction often yields multiple lexical variants of the same entity. For instance,
\textit{Section 302 IPC}, \textit{Sec. 302}, \textit{S. 302 of the Indian Penal Code}, and \textit{Section 302 of the Indian Penal Code} all refer to the same legal provision under Indian law.
Such variations fragment the graph and weaken semantic coherence, especially during aggregation. To overcome this, we perform entity canonicalization by identifying and grouping semantically equivalent entities.

We first obtain contextual embeddings of extracted entities using InLegalBERT\footnote{\url{https://huggingface.co/law-ai/InLegalBERT}}, a domain-specific model trained on Indian legal corpora. These embeddings are clustered agglomeratively using the cosine similarity, with a threshold of $0.95$ to merge only highly similar entities. To reduce redundancy while preserving fidelity, we retain all original mentions but assign the longest surface form in each cluster as the canonical name.

\end{enumerate}

\noindent 
Using the extracted triplets, we construct the knowledge graph corresponding to each segment $S_i^p \in J_i$, and denote it as  $\mathcal{G}i^p = (\mathcal{V}{i}^p, \mathcal{E}_{i}^p)$.

\begin{algorithm}[ht]
    \caption{PRecG Training Algorithm}
    \label{alg:judgment_representation_similarity}
    \KwIn{Training dataset $\mathcal{D}$ with $N$ triplets of the form $\{J_i, J_j, y_{ij}\}$}
    \KwOut{The set of learned parameters $\theta$}

    \textbf{Initialize} model parameters $\theta$ randomly\;

    \For{each triplet $\{J_i, J_j, y_{ij}\}$ in $\mathcal{D}$}{
        {\small{\tcp{Get representations for both judgments}}}
        \For{$J \in \{J_i, J_j\}$}{
            {\small{\tcp{Document Segmentation}}}
            $\mathcal{S} \leftarrow \text{Segment}(J)$\;
            {\small{\tcp{Segment-level KG Construction}}}
            \For{$S^p \in \mathcal{S}$}{
                $\mathcal{G}^p \leftarrow \text{ConstructGraph}(S^p)$\;
            }
            {\small{\tcp{Segment-level Representation Learning}}}
            \For{$\mathcal{G}^p$}{
                $\mathbf{s}^p \leftarrow \text{AttenPool}(\text{GATConv}(\mathcal{G}^p))$\;
            }
            {\small{\tcp{Judgment Representation Aggregation}}}
            $\mathbf{j} \leftarrow \text{TransfAggr}(\mathbf{s}^1, \dots, \mathbf{s}^k)$\;
            
            $\mathbf{j}_x \leftarrow \mathbf{j}, \quad x = i \text{ if } J = J_i \text{ else } j$\;

        }
        
         {\small{\tcp{Compute similarity and loss}}}
        $s_{ij} \leftarrow ComputeSimilarity(\mathbf{j}_i, \mathbf{j}_j)$\;
        $\mathcal{L}_\theta \leftarrow loss(y_{ij}, s_{ij})$\;

        {\small{\tcp{Update parameters}}}
        $\theta \leftarrow \text{Update}(\theta, \mathcal{L}_\theta)$\;
    }
    
    \Return $\theta$\;
\end{algorithm}

\subsection{Representation Learning via Graph Neural Networks}
\label{subsec:replearn}
This section describes the hierarchical process of learning the representation of case documents using a Graph Neural Network. First, each segment is encoded using the structural and semantic information captured in its knowledge graph. Next, the segment-level embeddings are aggregated using a transformer-based mechanism to obtain the representation of the entire document. Algorithm~\ref{alg:judgment_representation_similarity} summarizes the key steps, which are described below.

\begin{enumerate}[i)]
\item \textbf{Segment-level representation learning}:
Given the knowledge graph $\mathcal{G}_i^p = (\mathcal{V}_i^p, \mathcal{E}_i^p)$ of the $p^{th}$ segment ($S_i^p \in J_i$), our goal is to obtain a dense vector $\mathbf{s}_i^p$  that effectively encodes the structural and semantic information embedded in the segment.

Each node $v \in \mathcal{V}_i^p$, is initialized with an embedding ($\mathbf{h}_v^0 \in \mathbb{R}^d$) obtained from the associated text using InLegalBERT. Similarly, each edge $(u, v) \in \mathcal{E}_i^p$ is represented by a fixed embedding $\mathbf{e}_r \in \mathbb{R}^{d}$ derived from InLegalBERT.

To model the contextual interactions among entities, we employ a stack of $L$ Graph Attention Convolution (GATConv) layers~\cite{velickovic2018graph}. At each layer $\ell$, the attention coefficient $\alpha_{uv}^{(\ell)}$ quantifies the influence of node $u$ on node $v$, explicitly incorporating the relation embeddings $\mathbf{e}_r$ to encode the relational context within the graph.

\begin{equation}
\alpha_{uv}^{(\ell)} = \frac{\exp\left(\text{LeakyReLU}\left(\mathbf{a}^{\top} \left[ \mathbf{W}^{(\ell)} \mathbf{h}_u^{(\ell)} \, \| \, \mathbf{W}^{(\ell)} \mathbf{h}_v^{(\ell)} \, \| \, \mathbf{e}_r \right] \right)\right)}{\sum_{u' \in \mathcal{N}(v)} \exp\left(\text{LeakyReLU}\left(\mathbf{a}^{\top} \left[ \mathbf{W}^{(\ell)} \mathbf{h}_{u'}^{(\ell)} \, \| \, \mathbf{W}^{(\ell)} \mathbf{h}_v^{(\ell)} \, \| \, \mathbf{e}_{r'} \right] \right)\right)}
\end{equation}

\noindent Here, $\mathbf{W}^{(\ell)} \in \mathbb{R}^{d' \times d}$ a trainable transformation matrix, $\mathbf{a} \in \mathbb{R}^{3d'}$ a shared attention vector, and $\mathcal{N}(v)$ the neighborhood of node $v$. Notably, the edge embedding $\mathbf{e}_r$ remains fixed throughout training, supplying relational information without being updated.
During training, the representation of each node is updated by aggregating information from its neighbors, weighted by the corresponding attention coefficients. Additionally, a residual connection is incorporated to facilitate better gradient flow, followed by an ELU activation function (Eq. \ref{eqn:elu}).
We further apply layer normalization for stability (Eq. \ref{eqn:lnorm}).

\begin{equation}
\label{eqn:elu}
    \mathbf{h}_v^{(\ell+1)} = \text{ELU}\left( \sum_{u \in \mathcal{N}(v)} \alpha_{uv}^{(\ell)} \cdot \mathbf{W}^{(\ell)} \mathbf{h}_u^{(\ell)} + \mathbf{h}_v^{(\ell)} \right)
\end{equation}

\begin{equation} 
\label{eqn:lnorm}
    \hat{\mathbf{h}}_v^{(\ell+1)} = LayerNorm(\mathbf{h}_v^{(\ell+1)} ) 
\end{equation}

Thus, we obtain the embedding of node $v$ at layer $L$ as $\hat{\mathbf{h}}_v^{(L)}$.
The segment-level representation is then computed via attention-based pooling of all the node embeddings of $\mathcal{G}_i^p$.

\begin{equation}
\label{eqn:abp}
    \mathbf{s}_i^p = \sum_{v \in \mathcal{V}_i^p} \mathrm{attn}\big(\hat{\mathbf{h}}_v^{(L)}\big) \cdot \hat{\mathbf{h}}_v^{(L)}
\end{equation}

This pooled embedding $\mathbf{s}_i^p$ serves as the representation of segment $S_i^p$, forming the basis for document-level modeling.

%
\item \textbf{Document-level representation learning}:
We aggregate the segment embeddings $\{\mathbf{s}_i^1, \mathbf{s}_i^2,...., \mathbf{s}_i^k\}$ to obtain a \textit{unified} representation of the judgment $J_i$. Since different segments contribute differently to the semantic content of the judgment, we adopt a weighted aggregation mechanism that reflects the relative importance of each segment.
Accordingly, each segment $\mathbf{s}_i^p$ is projected to a shared vector space, parametrized by $\mathbf{W}^{(\text{a})} (\in \mathbb{R}^{d' \times d})$ and $\mathbf{b}^{(\text{a})} (\in \mathbb{R}^{d'})$.

\begin{equation}
    \mathbf{s'}_i^p = \mathbf{W}^{(\text{a})} \mathbf{s}_i^p + \mathbf{b}^{(\text{a})}
\end{equation}

To obtain a contextually enriched representation $\hat{\mathbf{s}}_i^p$, the projected vectors are passed through a transformer encoder, allowing each segment to attend to others.

\begin{equation}
    \hat{\mathbf{s}}_i^p = \sum_{m=1}^k \alpha_{m} \mathbf{s'}_i^m, \quad
    \alpha_{m} = \frac{\exp(\mathrm{score}(\mathbf{s'}_i^p, \mathbf{s'}_i^m))}{\sum_{j=1}^k \exp(\mathrm{score}(\mathbf{s'}_i^p, \mathbf{s'}_i^j))}
\end{equation}

Finally, mean pooling over these attention-weighted vectors produces the unified representation of the judgment.

\begin{equation}
    \mathbf{j}_i = \frac{1}{k} \sum_{p=1}^k \hat{\mathbf{s}}_i^p
\end{equation}

\end{enumerate}

\subsection{Training PRecG Model}
\label{subsec:simtrain}

Given the training dataset $\mathcal{D}$, containing $N$ instances, each of the form $\{J_i, J_j, y_{ij}\}$. The task is to model the similarity between the legal case documents $J_i$ and $J_j$, with $y_{ij} (\in [0,1])$ denoting the ground truth similarity score. A higher value indicates greater semantic similarity between $J_i$ and $J_j$. 

During training, each triplet is processed independently. Documents $J_i$ and $J_j$ are first segmented into rhetorical role-based units, from which entity-level graphs are constructed as described above. These graphs are encoded through GATConv layers to produce segment-level embeddings, which are then aggregated through a transformer to obtain document-level representations $\mathbf{j}_i$ and $\mathbf{j}_j$. The similarity $s_{ij}$ between these two judgments is then computed using the cosine similarity, which captures the degree of semantic alignment between the judgments. The training of the PRecG model incorporates both structural and contextual nuances, which are captured during the hierarchical encoding process. 
The model is trained using the focalMSE loss, which emphasizes hard-to-predict pairs by modulating the squared error term. The objective function is defined as:

\begin{equation}
    \mathcal{L}_\theta= \sum_{\substack{(J_i, J_j, y_{ij}) \in D }} \gamma \cdot \left(1 - \exp(-e_{ij}^2)\right) \cdot e_{ij}^2
\end{equation}

\noindent $\text{where},~e_{ij} =(y_{ij} - s_{ij})$ and $\theta$ denote the set of learnable model parameters. The deviation between the predicted similarity and ground truth $y_{ij}$ is backpropagated to update the parameters across all stages of the model, including segment encoding, graph propagation, and hierarchical aggregation.

\section{Experimental Setup}
\label{sec:experi}
In this section, we discuss the experimental setup for the evaluation of the PRecG model. We conducted experiments on a machine running Ubuntu 22.04.5 LTS, equipped with an Intel(R) Xeon(R) Gold 6226R @ 2.90GHz (64 cores) CPU, an NVIDIA RTX A6000 GPU (GA102 architecture), and 256 GB of RAM. We implement the model using PyTorch with CUDA support to leverage GPU acceleration. For LLM-based tasks, we use Ollama\footnote{\url{https://ollama.org/}} to run the LLaMA model locally and Neo4j\footnote{\url{https://neo4j.com/}} to store and query the knowledge graphs. 

\subsection{Research Questions}
We answer the following research questions through a carefully designed experimental study. 
\begin{enumerate}[RQ 1:]
    \item \textit{How effective is the learned judgment representation for similarity computation compared to the baselines?} 
    
    \noindent We address this research question by comparing our method against a diverse set of baselines for document similarity computation. We use a set of baselines that include $\mathcal{(\text{i})}$ statistical and contextualized representations of the judgment with no learning, $\mathcal{(\text{ii})}$ Contextualized paragraph-based representation with no learning, $\mathcal{(\text{iii})}$ Contextualized vector representation of judgment with learning, and $\mathcal{(\text{iv})}$ LLM-based similarity computation.

    \item \textit{Does rhetorical role-based segmentation improve model performance?} 
    
        \noindent The question is motivated by curiosity as to whether breaking down a judgment into semantically coherent segments provides richer contextual signals and enhances learning, as opposed to treating the document as a monolith. 
        We conduct an ablation study comparing two input strategies, viz., (i)  Knowledge graphs constructed from the segmented units of the judgment and (ii) a single knowledge graph built from the entire judgment text without segmentation, and compare their performance.
        
    \item \textit{Which segments of the judgment, when included, significantly affect model performance?}  
    
    \noindent In this ablation study, our aim is to identify the relative importance of the different rhetorical-role segments in the similarity computation between judgments. 
    
    \item \textit{How does the PRecG performance change when the knowledge graph is constructed using a traditional IE approach vs. an LLM-assisted approach?} 

    \noindent We compare the two approaches for knowledge graph construction (traditional IE vs. LLM-assisted) and compare the model performance under the two settings. The comparison helps us to understand whether the sophistication of LLM-enriched knowledge graphs provides any advantage over traditional approaches.

    \item \textit{What is the impact of entity and relationship initialization on the performance of the PRecG model?}

    \noindent We investigate how different initialization strategies for entity and relation representations influence the effectiveness of PRecG. In particular, we compare multiple legal domain–specific pretrained models.

\end{enumerate}

\subsection{Baselines}

The comprehensive representation of the judgment text is integral to automatic legal precedent retrieval. Accordingly, we include the following baseline models for judgment representation and similarity computation.

\begin{enumerate}[i)] 
    \item \textbf{Best Match 25 (BM25)}: 
    BM25 is a strong lexical baseline proposed in \cite{robertson2009probabilistic}.
    It enhances traditional statistics-based TF-IDF representation by incorporating term frequency saturation and document length normalization, making it particularly functional for information retrieval in the legal domain.
    
    \item \textbf{Paragraph Links Induced (PLI)}:
    PLI, introduced in \cite{kumar2014similarity}, captures the similarity of legal judgments using paragraphs as textual anchors. The method segments the judgment into paragraphs, encodes the paragraphs using the (pre-trained) BERT model, and computes the cosine similarity between paragraphs. Finally, paragraph links are established between two judgments based on a threshold. We compare the InLegalBERT and Doc2Vec models to encode the paragraphs. 
    The former is selected for its ability to understand the Indian legal context \cite{paul2022pretraining}, whereas the latter is chosen for its capability to capture document-level semantics \cite{lau2016empirical}.

     \item \textbf{InLegalBERT}:
    InLegalBERT, proposed in \cite{paul2022pretraining}, is a domain-specific language model pre-trained on Indian legal documents. It adapts the BERT architecture to capture the unique vocabulary, syntax, and semantic nuances of the Indian legal language. InLegalBERT has been used in several legal tasks, including named entity recognition \cite{bhandari2025transformer} and rhetorical role identification \cite{sheik2024enhancing}.
    We assess the representation capability of InLegalBERT for computing the similarity between judgments.

    \item \textbf{GAT Baseline}:
    We train a GATConv model \cite{velickovic2018graph} on the graph constructed from the judgments dataset to perform document-level similarity computation. We construct an unweighted and undirected graph that has judgments as nodes and edges between pairs of judgments in the dataset. Each judgment is encoded as a dense vector using pre-trained embeddings obtained from InLegalBERT. The GATConv model incorporates both the node features and the structural relationships among the documents through a self-attention mechanism. 
    \item \textbf{LLM Baseline}:
    We leverage the natural language understanding capabilities of LLMs to quantitatively assess how well the models can comprehend the legal intricacies of the judgments. We employ Llama3.1\footnote{Best open-source model at time of experimentation reported in this paper. URL: \url{https://ai.meta.com/blog/meta-llama-3.1/}} to assess the similarity between judgments and adopt the approach proposed in PromptResp\cite{zhuang2024promptreps}. This method constructs a dense and sparse representation of passages by prompting the LLM, extracting embeddings, and then computing document similarity using a hybrid similarity measure.
     
   The second LLM-based approach, outlined in \cite{halimi:2024}, decomposes the legal judgments into law points to obtain a high-level summary derived through a language model. 
    Subsequently, the law points of two judgments are passed into the LLM to compute a similarity score between them. 
\end{enumerate}

\subsection{Experimental Details}
We use the judgment similarity dataset introduced in \cite{bhattacharya-ipm22} for comparative evaluation. This human-annotated dataset comprises $190$ judgment pairs, each with an associated similarity score. The average similarity score between pairs is $0.49$, and the dataset covers judgment pairs from diverse legal domains. 

We conduct inductive evaluation for supervised models (PRecG and GAT baseline) by partitioning the dataset into five distinct train-test partitions, each with an $80:20$ split.  We report the average Mean Squared Error (MSE) and Mean Absolute Error (MAE), along with their standard deviations computed over the test sets. We execute the unsupervised baseline approaches on five test sets and report their average performance and standard deviation.

\begin{figure} 
    \centering

    \begin{subfigure}[b]{0.49\linewidth}
        \centering
        \includegraphics[width=0.9\linewidth]{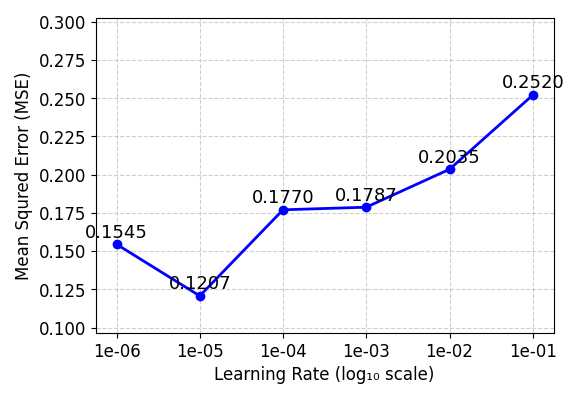}
        \caption{Learning rates $(\log_{10}~\text{scale})$}
        \label{fig:hpt_lr_mse}
    \end{subfigure}
    \hfill
    \begin{subfigure}[b]{0.49\linewidth}
        \centering
        \includegraphics[width=0.9\linewidth]{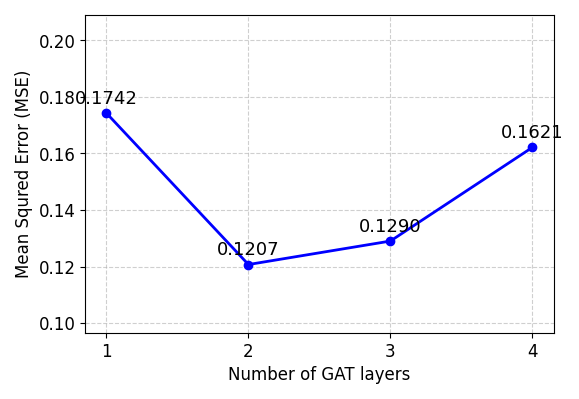}
        \caption{Number of GAT layers}
        \label{fig:hpt_gat_mse}
    \end{subfigure}
    \hfill
    \begin{subfigure}[b]{0.49\linewidth}
        \centering
        \includegraphics[width=0.9\linewidth]{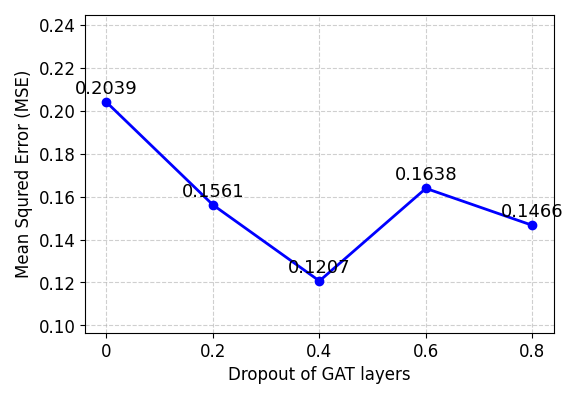}
        \caption{Dropout in GAT layers}
        \label{fig:hpt_do_mse}
    \end{subfigure}
    \hfill
    \begin{subfigure}[b]{0.49\linewidth}
        \centering
        \includegraphics[width=0.9\linewidth]{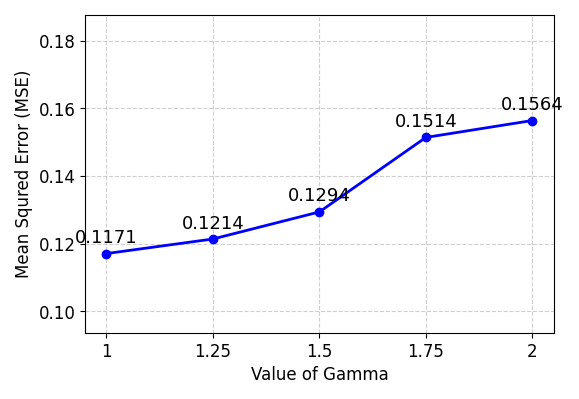}
        \caption{Gamma value of focalMSE}
        \label{fig:hpt_gma_mse}
    \end{subfigure}

    \caption{PRecG performance under different hyperparameter settings. Numbers in parentheses indicate the number of training epochs.}
    \label{fig:hpt_1x4}
\end{figure}

To ensure a fair comparison, we tune the hyperparameters for the GATConv and PRecG algorithms. The PRecG approach has four hyperparameters, namely, the learning rate, the number of GAT layers, the dropout ratio, and the gamma value of the focalMSE loss. For the learning rate, we test values in the range \{${10^{-1}, 10^{-2}, \dots, 10^{-6}}$\}, and find that $10^{-5}$ yields the best MSE. We vary the number of GAT layers from $1$ to $4$ and observe the best performance with $2$ layers. We search for the dropout ratio in the range $[0, 0.8]$ with a step size of $0.2$ and find that 0.4 yields the best accuracy. We test gamma values for the focalMSE loss in the range $\{1, 1.25, 1.5, 1.75, 2\} $and find that value $1$ yields the best result.
Figure~\ref{fig:hpt_1x4} shows the effect of hyperparameters on the PRecG model’s performance, measured by mean squared error (MSE). We tune the learning rate, number of GAT layers, and dropout for the GAT baseline within a similar range as the PRecG model. The results reported in the following sections use the best-performing hyperparameter settings.

\begin{table} 
\renewcommand{\arraystretch}{1.1}
\caption{Comparison of PRecG performance with baselines. }
\label{table:bl}
\centering
    \begin{tabular}{llcc}
        \toprule
        \textbf{} & \textbf{Model} & \textbf{MSE} & \textbf{MAE} \\
        \midrule
        \textbf{a} & \textbf{BM25} & 0.4807	$\pm$0.0120 & 0.6191 $\pm$0.0108  \\ \midrule   

        \multirow{2}{*}{\textbf{b}} & \textbf{PLI + Doc2Vec} & 0.7441 $\pm$0.0012 & 0.8267 $\pm$0.0007  \\ 
        & \textbf{PLI + InLegalBERT} & 0.3036 $\pm\text{0.0022}$ & 0.5144 $\pm$0.0019  \\ \midrule
        
        \textbf{c} & \textbf{InLegalBERT} & 0.3595 $\pm$0.0090 & 0.5204 $\pm$0.0071   \\ \midrule

        \textbf{d} & \textbf{GAT + InLegalBERT} & 0.8103 $\pm$0.1177 & 0.7634 $\pm$0.0374  \\ \midrule
        \multirow{2}{*}{\textbf{e}} & \textbf{PromptReps} & 0.5360 $\pm$0.0032 & 0.6992 $\pm$0.0020 \\ 
        & \textbf{Lawpoints} & 0.2560 $\pm$0.0075 & 0.5144 $\pm$0.0019  \\ \midrule 
        \textbf{f} & \textbf{PRecG} & 0.1171 $\pm$0.0093 & 0.2871 $\pm$0.0218   \\ 
        \bottomrule
    \end{tabular}
\end{table}

\section{Results and Discussion}
\label{sec:result}

In this section, we present the results of the experimental evaluation of our proposed PRecG model. The objective is to assess the model's ability to capture legal semantics and effectively retrieve similar judgments by answering the RQs mentioned in the previous section.

\subsection{Comparison with Baselines}

We compare the performance of the proposed PRecG model against five baselines that span a spectrum of document similarity approaches, ranging from simple statistical to sophisticated neural networks (Table \ref{table:bl}). It is clear from the table that GAT + InLegalBERT (row d) performs the worst. Despite incorporating the graph structure, its high MSE suggests that the model is overfitting, likely because of the limited training data. The PLI + Doc2Vec approach (row b) also performs inadequately, probably due to the generic nature of Doc2Vec embeddings, which struggle with the domain-specific language of legal judgments.
PromptReps (row e), which relies on LLMs through prompting, shows moderate performance. BM25 (row a), a statistical method, exhibits a performance comparable to the LLM-based baseline. However, its reliance on shallow lexical matching limits its ability to capture semantic nuances in legal language.

InLegalBERT, when used directly for dense representation of judgments (row c) performs better than all the baselines mentioned earlier, demonstrating the strength of domain-specific pretraining. 
When integrated with PLI (row b), it surpasses the vanilla usage, highlighting the value of explicitly modeling paragraph-level links.
Thus, PLI + InLegalBERT (row b) improves significantly over the Doc2Vec pairing, validating the impact of combining structure with strong domain-specific embeddings.
Lawpoints (row e) further improves upon LLM-based approaches, though it still lags behind PRecG. 

Overall, the PRecG model outperforms all the baselines with the least MSE and MAE. It goes beyond lexical surface matching, avoids reliance on generic embeddings, leverages domain-specific pretraining effectively, handles graph structure in a principled way, and captures the fine-grained nuances of legal judgments.

\subsection{Impact of Rhetorical Role-based Segmentation}

This experiment investigates the significance of the semantic segmentation step in the PRecG pipeline. We conduct an ablation study to compare model performance under two different configurations, one where the input to the pipeline has a single document-level knowledge graph, and the other in which it has a set of knowledge graphs constructed from rhetorical role-based segments.
This comparison allows us to isolate and assess the specific contribution of semantic segmentation to the overall model performance. As evident in Table \ref{tab:abla_1}, all evaluation metrics show a clear improvement when the input is segmented into rhetorical roles, rather than when the entire document is considered a monolith.

\begin{table} 
\renewcommand{\arraystretch}{1.1}
\caption{PRecG model performance with and without Rhetorical role-based semantic segmentation.}
\label{tab:abla_1}
    \centering
    \begin{tabular}{lcc}
        \toprule
        \textbf{Model}	& \textbf{MSE}	& \textbf{MAE} \\ 
        \midrule
        PRecG  & 0.1171	(\text{71.14\% ↑}) & 0.2871 (\text{49.38\% ↑})\\ 
        PRecG (w/o) & 0.4051	& 0.5672 \\
        \bottomrule
    \end{tabular}
\end{table}

Therefore, it is reasonable to conclude that semantic segmentation plays a critical role in improving the model's ability to accurately estimate similarity.

\subsection{Influence of Rhetorical Role-Based Segments on Model Performance}

Intuitively, the precedent judgments share partial, but meaningful overlap with the query judgment that might be within one or more semantic segments.  Since PRecG leverages segment-level representations to compute judgment-pair similarity, we evaluate the contribution of different segments through an ablation study, in which we incrementally incorporate knowledge graphs corresponding to different segment types. This stepwise inclusion allows us to isolate and quantify the contribution of segments to overall performance. In consultation with a legal expert, we adopted the order of inclusion of different segments as shown in Table~\ref{tab:abla_2}.

\begin{table}[h]
\renewcommand{\arraystretch}{1.1}
\caption{Performance of the PRecG model with incremental addition of different segment types.  F: Facts, S: Statute, A: Arguments, and R: Ratio Decidendi.}
\label{tab:abla_2}
\centering
\begin{tabular}{lcc}
\toprule
\textbf{Semantic Segment(s)} & \textbf{MSE} & \textbf{MAE} \\ 
\midrule
F & 0.1911 & 0.3566 \\ 
F + S & 0.1846 & 0.3609 \\ 
F + S + A & 0.1614 & 0.3303 \\ 
F + S + A + R & 0.1543  & 0.3165  \\ 
All & 0.1171  & 0.2871  \\ 
\bottomrule
\end{tabular}
\end{table}

The results presented in Table~\ref{tab:abla_2} clearly show that the incremental inclusion of segments consistently improves the performance of the model.
It is evident that each segment adds value individually, and their combination captures complementary aspects of legal judgment.

\subsection{Effect of Knowledge Graph Construction Method}

In this experiment, we investigate the efficacy of knowledge graphs constructed using large language models (LLMs) compared with knowledge graphs constructed using the classical information extraction (IE) method. For entity extraction, we employ the Indian Legal Named Entity Recognition (NER) model\cite{kalamkar2022named}, and for relation extraction, we use SpaCy’s Dependency Parser\footnote{\url{https://spacy.io/api/dependencyparser}}, which together extract knowledge triplets from the case documents.
Using the triplets extracted from this classical method, we construct knowledge graphs for the segments of the judgment and train the PRecG model.  Table~\ref{tab:abla_3} shows that the performance of the PRecG model with the classical approach to knowledge graph construction drops significantly compared to the LLM-based approach.

\begin{table}
\renewcommand{\arraystretch}{1.1}
\caption{Comparison of PRecG with the knowledge graph construction using LLM vs classical method for information extraction.}
\label{tab:abla_3}
    \centering
    \begin{tabular}{lcc}
        \toprule
        \textbf{Model}	& \textbf{MSE}	& \textbf{MAE}	 \\ 
        \midrule
        PRecG (LLM) & 0.1171(\text{91.26\% ↑})& 0.2871 (\text{65.62\% ↑}) \\ 
        PRecG (Classical)  & 1.3400	& 0.8350  \\ 
        \bottomrule 
    \end{tabular}
\end{table}

\subsection{Impact of Representation Initialization}

We examined the influence of different initialization strategies for entity and relationship embeddings on the performance of the PRecG model. Specifically, we compared three semantic encoding models: InLegalBERT~\cite{paul2022pretraining}, SAILER~\cite{li2023sailer}, and DELTA~\cite{li2025delta}, to evaluate how their contextual representations influence model performance. These two models have been selected for comparison because of their proficient legal judgment representation capabilities. This analysis aims to identify which pre-trained encoder yields the most effective foundation for learning the similarity between the judgments. 

\begin{table}[h]
\renewcommand{\arraystretch}{1.1}
\caption{Performance of PRecG with different encoding models.}
\label{tab:abla_5}
    \centering
    \begin{tabular}{lcc}
        \toprule
        \textbf{Model}	& \textbf{MSE}	& \textbf{MAE}	 \\ 
        \midrule
        PRecG + InLegalBERT & 0.1171 & 0.2871 \\ 
        PRecG + SAILER & 0.1698	& 0.3415  \\ 
        PRecG + DELTA & 0.1376	& 0.3275  \\ 
        \bottomrule 
    \end{tabular}
\end{table}

The results of this experiment are presented in Table~\ref{tab:abla_5}. It is evident that PRecG with initial embeddings from InLegalBERT outperforms both SAILER and DELTA embeddings. Although all three encoding models incorporate legal domain-specific pretraining, InLegalBERT provides more precise and contextually rich embeddings for the Indian legal context.

\section{Conclusion}
\label{sec:concl}

In this paper, we propose the PRecG model for automatic legal precedent retrieval. The model learns the hierarchical representation of the judgment, while taking into account of its semantic segments based on rhetorical roles. Our method significantly improves over existing approaches for automatic legal precedent retrieval by constructing knowledge graphs for each segment to encode critical legal details and relationships. The embeddings for all segments are then aggregated to yield a \textit{unified} representation of the judgment. This hierarchical representation learning framework leads to lower retrieval errors, as demonstrated by experiments on a benchmark Indian legal dataset. We further report detailed ablation studies to demonstrate the importance of rhetorical role-based segmentation of judgments and an advanced LLM-based approach for knowledge graph construction.



%

\bibliographystyle{unsrt}
\bibliography{bibtex/bib/refs}  

\end{document}